\documentclass{article}

\usepackage{microtype}
\usepackage{graphicx}
\usepackage{subfigure}
\usepackage{booktabs} 
\usepackage{multirow} 
\usepackage{pdfpages}
\usepackage{hyperref}
\usepackage{adjustbox}

\usepackage[sectionbib]{natbib}
\usepackage{bibunits}

\usepackage[accepted]{icml2023}

\usepackage{amsmath}
\usepackage{amssymb}
\usepackage{mathtools}
\usepackage{amsthm}
\usepackage{svg}
\usepackage[capitalize,noabbrev]{cleveref}
\usepackage{svg}

\usepackage{enumitem}

\theoremstyle{plain}

\theoremstyle{definition}

\theoremstyle{remark}

\usepackage[textsize=tiny]{todonotes}

\icmltitlerunning{Scalable Deep Learning for RNA Secondary Structure Prediction}

\defaultbibliographystyle{icml2023} 
\defaultbibliography{bib/rna,bib/rna_data,bib/rna_folding,bib/dl,bib/rna_design}

\begin{document}

\begin{bibunit}

\twocolumn[
\icmltitle{Scalable Deep Learning for RNA Secondary Structure Prediction}



\icmlsetsymbol{equal}{*}

\begin{icmlauthorlist}
\icmlauthor{Jörg K.H. Franke}{uni}
\icmlauthor{Frederic Runge}{uni}
\icmlauthor{Frank Hutter}{uni}
\end{icmlauthorlist}

\icmlaffiliation{uni}{Department of Computer Science, University of Freiburg, Freiburg, Germany}

\icmlcorrespondingauthor{Jörg Franke}{frankej@cs.uni-freiburg.de}

\icmlkeywords{Machine Learning, ICML}

\vskip 0.3in
]



\printAffiliationsAndNotice{}  

\begin{abstract}
The field of RNA secondary structure prediction has made significant progress with the adoption of deep learning techniques. In this work, we present the \emph{RNAformer}, a lean deep learning model using axial attention and recycling in the latent space. We gain performance improvements by designing the architecture for modeling the adjacency matrix directly in the latent space and by scaling the size of the model. Our approach achieves state-of-the-art performance on the popular TS0 benchmark dataset and even outperforms methods that use external information. Further, we show experimentally that the \emph{RNAformer} can learn a biophysical model of the RNA folding process.

\end{abstract}

\section{Introduction}
\label{intro}

RNA molecules play a central role in many cellular processes, including regulation of transcription, translation, epigenetics, or more general differentiation and development~\citep{morris2014rise}.
These functions strongly depend on the structure of the RNA, which is defined by the secondary structure that describes the intra-molecular basepair interactions, determined by the sequence of nucleotides. Also, the secondary structure can provide important insights into RNA behavior and guide the design of RNA-based therapeutics and nanomachines~\citep{kai2021structurally}.
Therefore, the accurate prediction of the secondary structure is very desirable and a significant problem in computational biology~\citep{bonnet2020designing}. 

Traditionally, the problem of secondary structure prediction is solved with dynamic programming approaches that minimize the free energy (MFE) of a structure, like the most widely used algorithm, RNAfold~\citep{hofacker_1994}. The optimization is based on thermodynamic parameters derived from UV melting experiments~\citep{szikszai2022deep}. 
More recently, deep-learning-based approaches have conquered the field, showing superior performance on benchmark datasets, and can further incorporate additional information e.g. embeddings from large-scale RNA sequence models~\citep{singh2019spotrna, chen2022rna_fm}.

We present in this work a deep learning architecture that outperforms other methods on a commonly used benchmark dataset, such as TS0 provided by \citet{singh2019spotrna}, without ensembling or making use of additional information.
Our performance improvements are mainly based on an axial attention Transformer-like architecture which has a potentially high inductive bias for the prediction of an adjacency matrix. In contrast to the conventional used CNNs, axial attention has a receptive field of the whole pair matrix at any time and does not need to build the receptive field by depth. Further, we gain improvement by recycling to simulate a larger depth and classical scaling in terms of more training data, model parameters, and longer training times. 

However, some work in the field recently raised concerns about the performance improvements of deep learning methods, questioning if the learned predictions are a result of similarities between training and test data, and if the algorithms really learn a biophysical model of the folding process~\citep{flamm2021caveats}. 
Since current datasets are typically curated with regard to sequence similarity only, the performance of models mainly assesses intra-family performance~\citep{szikszai2022deep}, while inter-family evaluations are rarely reported. 
Our suggestion is to show the capability to learn a biophysical model using sequences with predicted structures from the widely used, well-defined but simplified biophysical model RNAfold.
To this end, we build a dataset based on RNA family information from the Rfam~\citep{rfam_2003} database with structure predictions from RNAfold and demonstrate that our method is capable of learning the biophysical model of the folding process. 
Our main contributions are:

\begin{itemize}[label=$\bullet$, nosep, leftmargin=*, noitemsep, before={\vspace*{-0.25\baselineskip}}]
  \item We propose a novel architecture for RNA secondary structure prediction based on axial attention and recycling.
  \item We achieve state-of-the-art results on the commonly used benchmark dataset TS0 (Section \ref{sec:bpRNA}).
  \item We show that our method is capable of learning the underlying folding dynamics of an MFE model in an inter-family prediction setting (Section \ref{sec:Rfam}). 
\end{itemize}


\section{Background \& Related Work}
\label{rel_work}

Secondary structure prediction algorithms can be roughly divided into two classes: (1) \emph{de novo} prediction methods that seek to predict the structures directly from the nucleotide sequence and (2) \emph{homology modeling} methods that require a set of homologous RNA sequences for their predictions~\citep{singh2021spotrna2}, called an RNA family. 
Predictions can then be applied either within given families (intra-family predictions) or across different families (inter-family prediction).
\emph{De novo} prediction methods are typically preferred since the search for homologous sequences is time-consuming and often, there is no family information available for novel RNAs. 
Until recently, the field of \emph{de novo} RNA secondary structure prediction was dominated by Dynamic Programming (DP) approaches that either build on algorithms for predicting the MFE secondary structure~\citep{zuker_1981}, or algorithms to find the most likely structure (maximum expected accuracy). 
One disadvantage of these algorithms is that they are typically limited to the prediction of nested RNA secondary structures, i.e. they cannot predict Pseudoknots~\citep{staple2005pseudoknots} out-of-the-box, which are present in around 40\% of RNAs~\citep{Chen2020RNA}, overrepresented in functional important regions~\citep{staple2005pseudoknots} and known to assist folding into 3D structures~\citep{fechter2001}. 
Only recently, deep-learning-based approaches conquered the field, which benefit from making few assumptions on the underlying biophysical folding process, while not being restricted to only predict a subset of possible base pairs~\citep{singh2019spotrna}, and achieved state-of-the-art performance~\citep{chen2022rna_fm}.
We now briefly summarize some existing methods and refer the reader to more detailed related work in Appendix~\ref{app:related_work}.

\emph{RNAfold}~\citep{lorenz_2011} uses a DP approach for the prediction of MFE secondary structures. The version we use here is based on the energy parameters provided by the Turner nearest-neighbor model~\citep{turner2010nndb}.
\emph{SPOT-RNA}~\citep{singh2019spotrna} uses an ensemble of models with residual networks (ResNets)~\citep{he2016resnet}, bidirectional LSTM~\citep{schuster1997bidirectional}, and dilated convolution~\citep{yu2015multi} architectures. \emph{SPOT-RNA} was trained on a large set of intra-family RNA data for \emph{de novo} predictions on a newly proposed test set, TS0.
\emph{ProbTransformer}~\citep{franke2022probabilistic} uses a probabilistic enhancement of the Transformer architecture for intra-family predictions. The model is trained on a large set of available secondary structure data and evaluated on TS0.
\emph{RNA-FM}~\citep{chen2022rna_fm} uses sequence embeddings of an RNA foundation model that is trained on 23 million RNA sequences to perform intra-family predictions of RNA secondary structures in a downstream task. The foundation model consists of a large Transformer architecture, while the downstream model uses a ResNet32~\citep{he2016resnet}.

\begin{figure}[tb]
  \centering
  \includegraphics[width=0.9\columnwidth]{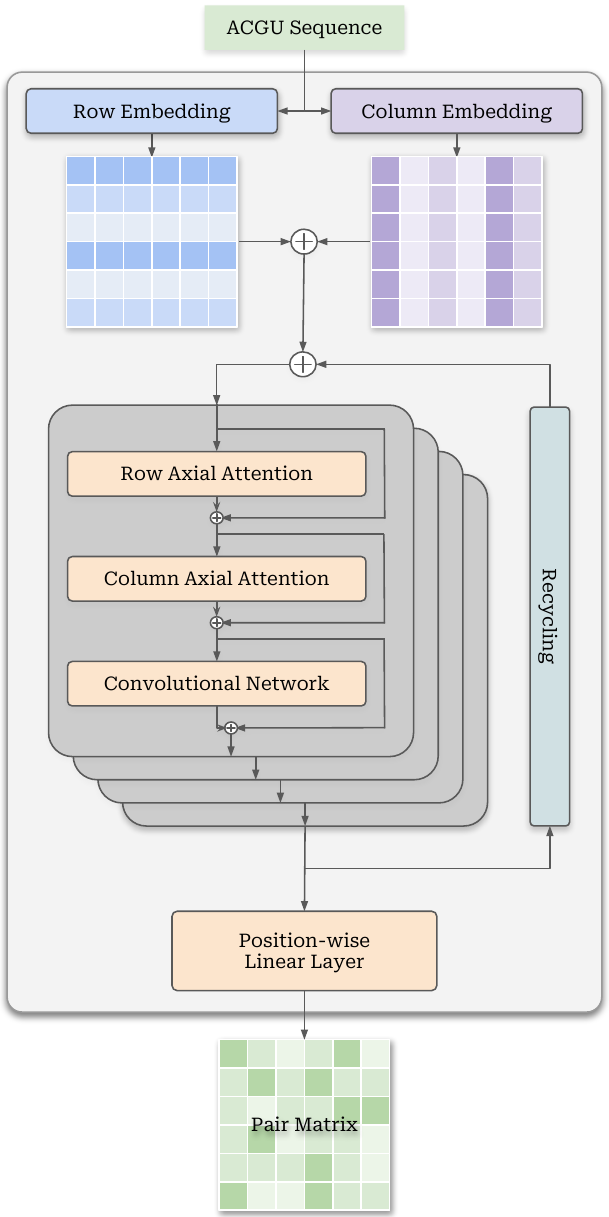}
  \caption{An overview of the \emph{RNAformer} architecture.}
  \label{fig:RNAformer}
\end{figure}
\section{RNAformer}
\label{RNAformer}

Our model architecture is inspired by AlphaFold~\citep{jumper2021highly}, which models a multi-sequence alignment and a pair matrix in the latent space and processes it with the use of axial attention~\citep{ho2019axial}. In our approach, which we dub \emph{RNAformer}\footnote{Source code and models: \href{https://github.com/automl/RNAformer}{github.com/automl/RNAformer}}, we simplify this architecture and only use axial attention for modeling a latent representation for the pairing between all nucleotides of the input RNA sequence. This construction leads to a potentially higher inductive bias since each layer adds some value to the latent representation of the adjacency matrix. To capture the dependency between the potential pairings we use two mechanisms: (1) \emph{axial attention} and a (2) \emph{convolutional layer}. Axial attention is a type of attention mechanism that captures dependencies between positions along a specific axis of the input data. In our case, we apply axial attention to the row and the column of the latent pairing matrix to create a dependency between all potential nucleotide pairings. To improve the modeling of local structures like stem-loops, we use a convolutional neural network with a kernel size of three instead of the position-wise feed-forward layer from the vanilla Transformer~\citep{attention2017vaswani}.  

\emph{RNAformer} embeds the RNA sequence with a linear layer twice and broadcasts them, one for a row-wise and one for a column-wise representation before we add them to the initial latent representation. Now we apply multiple Transformer-like blocks, each consisting of a row-wise axial attention, a  column-wise axial attention, and a two-layer convolutional network. Lastly, we apply a linear layer and output the paring matrix of the secondary structure directly. 
Similar to AlphaFold, we apply \emph{recycling} of the processed latent space to artificially increase the model depth and allow the model to reprocess and correct its own predictions. Therefore, we pass the latent representation multiple times through the block without gradient and calculate gradients only for the last recycle iteration. 
We apply dropout, pre-norm, and residual connections to all layers except the embedding and generator layers. For loss calculation, we masked 50\% of the unpaired entries in the adjacency matrix before calculating the mean cross-entropy loss. This helps to increase the learning signal in the heavily imbalanced adjacency matrix.  Refer to Figure~\ref{fig:RNAformer} for an overview of our architecture.

\section{Experiments}

We evaluate the performance of our model in two settings. First, we evaluate the intra-family prediction capability based on the bpRNA dataset. Secondly, we assess the performance on inter-family predictions, as well as investigate the learning of a biophysical model by training the \emph{RNAformer} on a dataset derived from Rfam database and the generated target secondary structures with RNAfold. 

\subsection{bpRNA Experiment} 
\label{sec:bpRNA}

\paragraph{Data curation} In order to generate a training dataset for intra-family predictions, we first collect a large data corpus from the following public sources: the bpRNA-1m~\citep{danaee2018bprna}, the ArchiveII~\citep{sloma_2016} and RNAStrAlign~\citep{tan2017stralign} dataset provided by \citet{Chen2020RNA}, all data from RNA-Strand~\citep{andronescu2008rna}, as well as all RNA containing data from PDB. 
Secondary structures for PDB samples were derived from the 3D structure information using DSSR~\citep{lu2015dssr}.
After removing duplicates we use the exact same protocol as \citet{singh2019spotrna} to remove sequence similarities while we replace the training set TR0 with our own data. In particular, we apply a 80\% similarity cutoff between the sequences using CD-HIT~\citep{fu2012cdhit} and a homology search using BLASTN~\citep{altschul1997gapped} with a large e-value of 10, to further reject sequences from our training set that show homologies with the respective test sets.
Most DL methods use the TS0 dataset for evaluations. 
However, similar to ~\citet{franke2022probabilistic}, we did not cluster the training, validation, and test data internally to learn from the data diversity.

\begin{table}[tb]
\centering
\caption{The mean performance of three runs with different random seeds in comparison on the TS0 benchmark dataset. We evaluated all competitors based on their open-sourced models and will publish our evaluation script with the model and code.\\ }
\label{tbl:ts0}
\begin{tabular}{@{}lrrr@{}}
\toprule
\multirow{2}{*}{Model} & \multicolumn{3}{c}{TS0} \\ \cmidrule(l){2-4}
      &  F1 Score & MCC & Solved \\ \midrule
RNAformer $32M +\circlearrowleft$ & \textbf{0.728} &  \textbf{0.733} & \textbf{17.2}\% \\
RNAformer $32M$ & 0.717 & 0.727  & 16.6\% \\
RNAformer $8M$ & 0.708  & 0.716 & 14.4\% \\
RNAformer $2M$ & 0.677 & 0.684 & 11.4\% \\
RNAformer $0.5M$ & 0.644 & 0.653  & 8.7\% \\ \midrule
RNA-FM & 0.667 & 0.671 & 10.4\% \\
ProbTransformer & 0.625 &  & 11.8\% \\
SPOT-RNA & 0.597 & 0.597  & 0.05\% \\
RNAfold & 0.492 & 0.499  & 0.8\% \\
\bottomrule
\end{tabular}
\end{table}

\paragraph{Model \& Training Setup} We evaluate the \emph{RNAformer} in a setup with $6$ blocks and with different latent dimensions of 32, 64, 128, and 256, resulting in total parameter counts of roughly 0.5M, 2M, 8M, and 32M parameters, respectively. We applied recycling ($\circlearrowleft$) with 6 iterations to the largest model and sample the number of recycle iterations during the training uniformly from 2 to 6. We trained all models on 8 GPUs with a batch size of 500 tokens per GPU and a maximum sequence length of 500, for 50k steps. This limit is mainly due to the large memory footprint of the two-dimensional latent space, however, we note that the same cutoff was also applied in previous work~\citep{singh2019spotrna}. For optimization, we used AdamW \citep{loshchilov2019decoupled} learning rate warm-up, a cosine learning rate decay, weight decay, and gradient clipping. Refer to Appendix \ref{app:training} for all hyperparameter values.

\paragraph{Results} We compared \emph{RNAformer} to the models in the related work and present the results in Table \ref{tbl:ts0}. For a more comprehensive comparison refer to Appendix \ref{app:experiments}. Our largest model with 32M parameters with the use of recycling achieves a new state-of-the-art result on the TS0 benchmark set. We solve 17.2\% of the sequences completely without any mistakes. The recycling ($\circlearrowleft$) leads to a performance gain of $\sim1\%$ and a steady increase of the parameter count from 0.5M to 32M also leads to a steady performance increase. This shows that we gain performance from over-parameterization and could indicate that the inductive bias induced by the architecture is beneficial for this task. 

\subsection{Rfam Experiment} 
\label{sec:Rfam}

\paragraph{Data curation} To evaluate the performance on inter-family predictions, as well as investigate the learning of a biophysical model, we derive a training dataset from families of the Rfam database version 14.9 \citep{kalvari2020rfam}. 
We first select all families with a covariance model with maximum \emph{CLEN} of $\leq$ 500 and sample a large set of sequences for each family from the covariance models using Infernal~\citep{nawrocki2013infernal}. 
We then build a large set with two third sequences from families with \emph{CLEN} $\leq$ 200 and one-third of sequences from the families with \emph{CLEN} $>$ 200 to increase the number of families further.
We randomly select 25 and 30 families from this set for validation and testing, respectively, and leaf all samples from other families for training.
All sequences are folded using \emph{RNAfold}~\citep{lorenz_2011}.
We apply a length cutoff at 200 nucleotides since we expect \emph{RNAfold} predictions to be more reliable for sequences below this threshold, to save computational costs, and since all datasets of experimentally derived RNA structures from the literature show a maximum sequence length below 200 nucleotides.
\citet{singh2021spotrna2} created a test set, TS-hard, in an inter-family manner similar to the data pipelines used by the Rfam database for RNA family assignments. 
We follow this pipeline to remove similar sequences between our training data and the validation- and test sets provided by \citet{singh2021spotrna2} using CD-HIT and BLASTN as described before. 
We then build an MSA of all sequences in TS-hard with BLASTN at an e-value of $0.1$ using NCBI's nt database as a reference and build covariance models from the MSAs using Infernal. However, while \citet{singh2021spotrna2} used \emph{SPOT-RNA} for predictions of the consensus structures of the MSA, which appears inappropriate since the method was built for \emph{de-novo} predictions, we use \emph{LocARNA-P}~\citep{will2012locarna}, a commonly used tool to build MSAs based on sequence and structure-based alignments. The covariance models were then used to remove all sequences from the training data, using an e-value threshold of $0.1$. We use this dataset to learn the underlying biophysical model of \emph{RNAfold}, evaluated on the Rfam test data, and for evaluations on TS-hard.
Again we avoid clustering the datasets internally to keep structural diversity.
All datasets are described in more detail in Table~\ref{tbl:datasets} in Appendix~\ref{app:data}.

\paragraph{Model \& Training Setup} We used the same setup as in the first experiment with the difference of a maximum sequence length of 200 tokens, a batch size of 600 tokens per GPU, and a training time of 100k steps. 

\begin{table}[tb]
\centering
\caption{We train different sizes of our model on the Rfam dataset on three different random seeds and report the mean performance.  }
\label{tbl:rfam}
\begin{tabular}{@{}lrr|r}
\toprule
\multirow{2}{*}{Model} & \multicolumn{2}{c|}{Rfam TS} & TS-hard \\  \cmidrule(l){2-4}
      &  F1 Score & Solved &  F1 Score \\ \midrule
RNAformer $32M +\circlearrowleft$ & \textbf{0.967} & \textbf{84.5}\% & \textbf{0.651} \\
RNAformer $32M$ & 0.936 & 65.2\% & 0.642 \\
RNAformer $8M$ & 0.925 & 60.0\% & 0.639 \\
RNAformer $2M$ & 0.870 & 37.2\% & 0.625 \\ \midrule
RNAfold &  &  & 0.636 \\
\bottomrule
\end{tabular}
\end{table}

\paragraph{Results}

As shown in Table \ref{tbl:rfam}, we can replicate the RNAfold algorithm increasingly better with growing model size. Our largest model achieves a mean F1 score of 94.8 on the test set and predicts 76.3\% of the structures entirely correct. This result suggests that the \emph{RNAformer} can learn the underlying biophysical model of the folding process. We observe similar results regarding scaling for the TS-hard dataset, where F1 scores increase with model size, resulting in a similar performance as RNAfold, which further supports our observation on the Rfam dataset. Interestingly, our larger models even slightly outperform RNAfold on TS-hard. However, these results require further investigations and a closer look at what the \emph{RNAformer} layers models in detail, before we speculate about whether these results originate from the \emph{RNAformer} architecture, or simply from slight deviations from the learned biophysical model.

\section{Conclusion \& Future Work}

We introduced a new architecture for RNA secondary structure prediction and showed state-of-the-art performance on the TS0 benchmark set. The gain in performance is based on axial attention, a recycling of the latent space, and a larger dataset based on the same similarity criteria as used in related work. We also trained the \emph{RNAformer} on a dataset derived from the Rfam database with RNAfold prediction to demonstrate that we can learn a biophysical model like RNAfold. The downside of our approach is a large memory footprint.
Our approach could be further improved by the usage of additional information like MSA \citep{singh2021spotrna2} or language embeddings with additional text information. We could also improve the architecture and enhance it with a probabilistic layer to capture ambiguities \citep{franke2022probabilistic} or scale it even further. Another way to improve or adapt our model is finetuning, which is heavily used for large language models and could be applicable to fine-tuning high-quality data. 
However, besides methodological improvements, more effort in the generation and collection of high-quality data is required to achieve  accurate predictions of RNA structures with deep learning.

\section*{Acknowledgements}

This research was funded by the Deutsche Forschungsgemeinschaft (DFG, German Research Foundation) under grant number 417962828.


\putbib
\end{bibunit}

\newpage
\appendix
\onecolumn

\begin{bibunit}

\begin{center}
  \vspace*{2em} 
  \Large\textbf{Appendix}
  \vspace{2em}
\end{center}

\section{Training Details}
\label{app:training}

\begin{table}[htbp]
\centering
\begin{tabular}{p{0.15\textwidth}p{0.25\textwidth}p{0.3\textwidth}}
\toprule
Group & Parameter & Value \\
\midrule
\multirow{5}{*}{Training} & accelerator & GPU \\
& devices & 8 \\
& gradient\_clip\_val & 1.0 \\
& max\_steps & 50000 (100000) \\
& seed & 1 / 2 / 3 \\
\midrule
\multirow{9}{*}{Optimizer} & optimizer & AdamW \\
& learning\_rate & 0.001 \\
& weight\_decay & 0.1 \\
& betas & [0.9, 0.98] \\
& eps & 1.0e-09 \\
& adam\_w\_mode & true \\
& num\_warmup\_steps & 2000 \\
& decay\_factor & 0.01 \\
& schedule & cosine annealing \\
\midrule
\multirow{16}{*}{Model} & vocab\_size & 5 \\
& max\_len & 500 (200) \\
& model\_dim & 256 / 128 / 64 / 32 \\
& n\_layers & 6 \\
& num\_head & 4 / 2 / 1 / 1 \\
& ff\_kernel & 3 \\
& cycling & 6 \\
& resi\_dropout & 0.1 \\
& embed\_dropout & 0.1 \\
& relative position encoding & True \\
& ln\_eps & 1e-5 \\
& softmax\_scale & True \\
& key\_dim\_scaler & True \\
& flash\_attn & True \\
& initializer\_range & 0.02 \\
\midrule
\multirow{8}{*}{Data} & dataset & bpRNA (Rfam) \\
& random\_ignore\_mat & 0.5 \\
& num\_cpu\_worker & 32 \\
& num\_gpu\_worker & 8 \\
& min\_len & 10 \\
& max\_len & 500 (200) \\
& batch\_token\_size & 500 (600) \\
& shuffle\_pool\_size & 100 \\
\bottomrule
\end{tabular}
\caption{The hyperparameters of the \emph{RNAformer} training.}
\label{tab:hyperparameters}
\end{table}

\newpage

\section{Related Work}
\label{app:related_work}

As described in Section~\ref{rel_work}, RNA secondary structure prediction was previously dominated by dynamic programming approaches the either optimize for MFE or maximum expected accuracy (MEA) predictions. 
The runtime of these approaches in $\mathcal{O}\left(n^3\right)$.
However, linear time approximations have been proposed~\cite{huang2019linearfold}.
Besides runtime, the major disadvantage of these algorithms is that they are typically limited to the prediction of nested RNA secondary structures, which strongly limits their accuracy~\citep{szikszai2022deep}.
Some work, however, used heuristic approaches to overcome this issue, again at the price of runtime~\citep{theis2010prediction, sato2011ipknot}.

In this regard, deep learning approaches have strong advantages, especially when modeling the RNA secondary structure as an adjacency matrix, where all types of pairs and pseudoknots are represented identically.
We now discuss existing deep learning approaches in more detail.

\emph{SPOT-RNA}~\cite{singh2019spotrna} was the first algorithm using deep neural networks for end-to-end prediction of RNA secondary structures, using an ensemble of models with residual networks (ResNets)~\cite{he2016resnet}, bidirectional LSTM-~\citep{hochreiter1997lstm} (BiLSTMs)~\citep{schuster1997bidirectional}, and dilated convolution~\citep{yu2015multi} architectures. \emph{SPOT-RNA} was trained on a large set of intra-family RNA data for \emph{de novo} predictions on TS0, and further fine-tuned on a small set of experimentally-derived RNA structures, for predictions including tertiary interactions. However, the performance for these types of base pairs was rather poor and the currently available version of the algorithm excludes tertiary interactions from its outputs.

\emph{E2efold}~\citep{Chen2020RNA} uses a Transformer encoder architecture for \emph{de novo} prediction of RNA secondary structures. The algorithm was trained on a dataset of homologous RNAs and showed strongly reduced performance across evaluation in multiple other publications~\citep{sato2021mxfold2, fu2022ufold}, which indicates strong overfitting. We use the same data as the respective work for evaluations and thus exclude \emph{E2efold} from our evaluations. 

\emph{MXFold2}~\citep{sato2021mxfold2} seeks to learn the scoring function for a subsequent DP algorithm using a CNN/BiLSTM architecture. The network is trained to predict scores close to a set of thermodynamic parameters. In contrast to the previously described methods, \emph{MXFold2} is restricted to predicting a limited set of base pairs due to the DP algorithm.

\emph{UFold}~\citep{fu2022ufold} employs a UNet~\citep{ronneberger2015unet} architecture for \emph{de novo} secondary structure prediction, additionally reporting results for predictions on data that contains tertiary interactions after fine-tuning the model.
In \emph{UFold} an RNA sequence is an image of all possible base-pairing maps and an additional map for pair probabilities, represented as square matrices.

\emph{SPOT-RNA2}~\citep{singh2021spotrna2} is a \emph{homology modeling} method that incorporates MSA features as well as sequence profiles (PSSM) and features derived from direct coupling analysis (DCA) for the prediction of RNA secondary structures. Similar to \emph{SPOT-RNA}, predictions are based on an ensemble of models but using dilated convolutions only. Since \emph{SPOT-RNA2}'s predictions are based on evolutionary features and homologous sequence information, the predictions can be considered intra-family wise independent of the curation of the dataset since homologies between the evolutionary information and the training or test sets were not explicitly excluded during evaluations. Nevertheless, we use the carefully designed test set, TS-hard, proposed by \citet{singh2021spotrna2} for our evaluations on inter-family predictions as described in Section~\ref{app:data}.

\emph{ProbTransformer}~\citep{franke2022probabilistic} uses a probabilistic enhancement for either an encoder or decoder transformer architecture for intra-family predictions. The model is trained on a large set of available secondary structure data and evaluated on TS0. By learning a hierarchical joint distribution in the latent, the ProbTransformer is the first learning algorithm that is capable of sampling different structures of this latent distribution, which was shown by reconstructing structure ensembles of a distinct dataset with multiple structures for a given input sequence.

\emph{RNA-FM}~\citep{chen2022rna_fm} uses sequence embeddings of an RNA foundation model that is trained on 23 million RNA sequences from 800000 species to perform intra-family predictions of RNA secondary structures in a downstream task. The foundation model consists of a 12-layer transformer architecture, while the downstream models use a ResNet32 architecture.

\emph{REDfold}~\citep{chen2023redfold} uses a residual encoder-decoder architecture inspired by the UNet architecture of UFold. Interestingly, the model input is a $146\times L\times L$ tensor, representing square matrices of all possible base pairs (10 combinations for dinucleotide pairs) and tetranucleotide combinations (136 combinations) without considering their order.
The model is trained on highly homogeneous data, reporting strong performance on 4-fold cross-validation experiments, but also reporting strong results when considering sequence similarity. However, when we evaluated REDfold on TS0, we did not observe the same performance (see Table~\ref{tbl:ts0_app}). Together with the results on unseen families provided by~\citet{chen2023redfold}, this might indicate potential overfitting. 

We note that there are other methods we do not consider here because they either showed inferior performance to methods we compare against~\citep{zhang2019new, rezaur2019learning, saman2022rna, wayment2022rna} or because their source code is not publicly available~\citep{jungrtfold}.


\section{Data}
\label{app:data}

\begin{table}[ht]
\centering
\begin{tabular}{@{}lrrrr@{}}
\toprule
Dataset & \# Samples & Min -- Max Length & Mean Length & \# Families \\
\midrule
TS-hard          & 28                  &         34 -- 189          &  65.6                &         --           \\
Rfam Test        & 3344                &         37 -- 182          &  79.4                &         30           \\
Rfam Valid       & 2727                &         34 -- 160          &  80.2                &         25           \\
Rfam Train       & 410408              &         22 -- 200          &  95.2                &        3796          \\
\midrule
TS0              & 1305                &         22 -- 499          &  136.1               &         --           \\
VL0              & 1291                &         33 -- 497          &  132.1               &         --           \\
bpRNA Train      & 40836               &         13 -- 500          &  123.0               &         --           \\
\bottomrule
\end{tabular}
\caption{Dataset overview.}
\label{tbl:datasets}
\end{table}

\section{Experiments}
\label{app:experiments}

\begin{table}[h]
\centering
\begin{tabular}{@{}lrrrr@{}}
\toprule
\multirow{3}{*}{Model} & \multicolumn{4}{c}{TS0} \\ \cmidrule(l){2-5}
      &  \multicolumn{2}{c}{F1 Score} & \multicolumn{2}{c}{Solved} \\ \cmidrule(lr){2-3} \cmidrule(lr){4-5}
      &  mean & std & mean & std \\ \midrule
RNAformer $32M +\circlearrowleft$ & \textbf{0.728} & 0.002 &  \textbf{17.2\%} & 0.002 \\
RNAformer $32M$ & 0.717  & 0.002 & 16.6\% & 0.001 \\
RNAformer $8M$ & 0.708  & 0.001 & 14.4\% & 0.003 \\
RNAformer $2M$ & 0.677  & 0.005 & 11.4\% & 0.001 \\
RNAformer $0.5M$ & 0.644  & 0.003 & 8.7\% & 0.002 \\ \midrule
RNA-FM* & \multicolumn{2}{c}{0.667} & \multicolumn{2}{c}{10.4\%} \\
ProbTransformer & \multicolumn{2}{c}{0.625} & \multicolumn{2}{c}{11.8\%} \\
SPOT-RNA & \multicolumn{2}{c}{0.597} & \multicolumn{2}{c}{0.05\%} \\
MXFold2 & \multicolumn{2}{c}{0.550} & \multicolumn{2}{c}{1.4\%} \\
UFold & \multicolumn{2}{c}{0.588} & \multicolumn{2}{c}{3.8\%} \\
RNAfold & \multicolumn{2}{c}{0.492} & \multicolumn{2}{c}{0.8\%} \\
LinearFold-C & \multicolumn{2}{c}{0.509} & \multicolumn{2}{c}{1.2\%} \\
LinearFold-V & \multicolumn{2}{c}{0.493} & \multicolumn{2}{c}{0.8\%} \\
RNAStructure & \multicolumn{2}{c}{0.490} & \multicolumn{2}{c}{0.6\%} \\
pKiss & \multicolumn{2}{c}{0.450} & \multicolumn{2}{c}{0.3\%} \\
CONTRAfold & \multicolumn{2}{c}{0.522} & \multicolumn{2}{c}{0.8\%} \\
IpKnot & \multicolumn{2}{c}{0.504} & \multicolumn{2}{c}{0.4\%} \\
REDfold & \multicolumn{2}{c}{0.475} & \multicolumn{2}{c}{2.2\%} \\ \bottomrule
\end{tabular}
\caption{Performance comparison on the TS0 benchmark dataset. We report the mean and standard derivation of the performance of three RNAformer runs with different random seeds. *The number differs from their publication since we used their open-sourced model and our evaluation script which will be publicly available upon acceptance. We note, however, that the \emph{RNAformer} also achieves a higher F1 score than reported in the publication of RNA-FM.}
\label{tbl:ts0_app}
\end{table}

\newpage
\twocolumn

\putbib
\end{bibunit}

\end{document}